\documentclass[letterpaper]{article} 
\usepackage{main}  
\usepackage{times}  
\usepackage{helvet} 
\usepackage{courier}  
\usepackage[hyphens]{url}  
\usepackage{graphicx} 
\usepackage[compress]{natbib}  
\usepackage{caption} 
\usepackage{fullpage}
\usepackage{hyperref}

\title{A Short Survey of Systematic Generalization}
\author{
    Yuanpeng Li\thanks{\texttt{yuanpeng16@gmail.com}}
}

\begin{document}

\maketitle

\begin{abstract}
This survey includes systematic generalization and a history of how machine learning addresses it.
We aim to summarize and organize the related information of both conventional and recent improvements.
We first look at the definition of systematic generalization, then introduce Classicist and Connectionist.
We then discuss different types of Connectionists and how they approach the generalization.
Two crucial problems of variable binding and causality are discussed.
We look into systematic generalization in language, vision, and VQA fields.
Recent improvements from different aspects are discussed.
Systematic generalization has a long history in artificial intelligence.
We could cover only a small portion of many contributions.
We hope this paper provides a background and is beneficial for discoveries in future work.
\end{abstract}

\section{Introduction}
Artificial intelligence with deep learning has rapid improvement in recent years.
As it addresses many problems, an old question of systematic generalization~\cite{fodor1988connectionism,lake2018generalization} returns to receive focus.
Systematic generalization requires correctly addressing unseen samples by recombining seen ones.
For example, a model trained with blue rectangles and green triangles predicts blue triangles.
Though it is straightforward for humans, it is still challenging for deep learning models.
It has a long history and many recent related works.
This survey summarizes and organizes the information into a series of subtopics.
The initial ones mainly focus on historical perspectives, and the latter discuss recent works.
We will go through them in this introduction and look at their details in the following sections.

The fast growth of deep learning has addressed many i.i.d. problems in artificial intelligence.
On the other hand, systematic generalization is an out-of-distribution (o.o.d.) generalization with disjoint training and test domains.
It provides the ability for fast learning and creation.
They are essential for more human-like intelligence, which current machine learning does not achieve (Section~\ref{sec:systematic_generalization}).

Artificial intelligence has Classicist and Connectionist approaches, and they have complementary advantages.
Connectionist~\cite{feldman1982connectionist} is good at i.i.d. generalization but not at systematic generalization~\cite{fodor1988connectionism,marcus1998rethinking}.
Classicist with symbol processing is the opposite.
A neural network originates from the Connectionist model and is still weak at such generalization (Section~\ref{sec:classicist_and_connectionist}).

Three types of Connectionist are explored to facilitate the advantages of Classicist and Connectionist.
Eliminative Connectionist does not use symbolic processing.
Hybrid Connectionist involves both Connectionist and symbolic processing.
Implementational Connectionist uses Connectionism to implement symbolic processing (Section~\ref{sec:types_of_connectionist}).

Variable binding is a Connectionist topic closely related to systematic generalization.
A variable is a placeholder, and it can be replaced with values.
It decouples the manipulation of a variable and its value, so it generalizes to their combinations (Section~\ref{sec:variable_binding}).

Causality is also a related topic.
It uses do-calculus with intervention and enables studying probabilities of counterfactual events.
Systematic generalization is such a counterfactual event (Section~\ref{sec:causality}).

Systematic generalization problems are widely encountered in different fields.
We mainly focus on language, vision, and visual question answering (VQA).
Many datasets are designed in these fields.
Language has historically been more studied since sentences are more straightforward to process than images.
The recent refocus on systematic generalization also started from language tasks~\cite{lake2018generalization} (Section~\ref{sec:language_vision_vqa}).

Many deep learning approaches have been recently proposed for systematic generalization, such as disentangled representation learning, meta-learning, attention mechanism, modular architecture, specialized architectures, and data augmentation (Section~\ref{sec:recent_improvements}).

We cover the history of systematic generalization in artificial intelligence and summarize the recent development after the wide use of deep learning.
We hope this survey is helpful as background information for potential future research.
The following sections have more detailed discussions for each subtopic mentioned above.

\section{Systematic Generalization}
\label{sec:systematic_generalization}
A fundamental property of artificial intelligence is generalization, where a trained model appropriately addresses unseen test samples.
Many problems adopt the i.i.d. assumption, where training and test samples are independently drawn from the identical distribution (i.i.d. generalization).
On the other hand, test distribution can be different from training distribution, and they may have disjoint input domains or support.
It means that the test samples have zero probability in training distribution (o.o.d. generalization).

Systematic generalization is an o.o.d. generalization.
Systematicity is a property where the ability to produce or understand some sentences (or objects in general) is intrinsically connected to that ability for others~\cite{fodor1988connectionism}.
It usually uses factors of variation~\cite{bengio2013representation} for recombination.
For example, a model trained with blue rectangles and green triangles predicts blue triangles.
A systematic generalization is often called a compositional generalization, mainly in language domains.

Systematic generalization is considered the "Great Move" of evolution, developed to process an increasing amount and diversity of information from the environment; for example, humans can recognize new spatial relationships of seen objects~\cite{10.5555/86564}.
It is also related to the evolution of the prefrontal cortex~\cite{1995Relational}.
Cognitive scientists see such generalizations as central for an organism to view the world~\cite{2009Memory}.
A book~\cite{calvo2014architecture} contains works mainly from cognitive science perspectives.

It has been discussed that commonsense is critical~\cite{1959Program,1986CYC} for systematic generalization.
The discussions also seek general prior knowledge for systematic generalization, e.g., Consciousness Prior~\cite{bengio2017consciousness}.
Recently, different types of inductive bias~\cite{goyal2020inductive} were summarized.
%
There are different ways to categorize systematicity.
Three levels of systematicity, from weak to strong, are defined~\cite{hadley1992compositionality}.
More precisely, six levels of systematicity are defined~\cite{Niklasson94canconnectionist}. 
Recently, five types of tests are summarized~\cite{ijcai2020-708}.

The development of deep learning reliably addresses many i.i.d. problems, so it is also encouraged to address systematic generalization.
It relates to many areas, including
reasoning~\cite{talmor2020leap},
continual learning~\cite{Li2020Compositional},
zero-shot learning~\cite{sylvain2020locality}, and
language inference~\cite{geiger-etal-2019-posing}.
GFlowNets~\cite{bengio2021flow} generate informative samples to address systematic generalization for active learning and exploration in reinforcement learning.

\section{Classicist and Connectionist}
\label{sec:classicist_and_connectionist}
Artificial Intelligence has been developed in two approaches: Classicists and Connectionists.
They are discussed for human cognition and refer to approaches in artificial intelligence.

Classicist~\cite{fodor1975the,pylyshyn1980computation} refers to the computing operations on symbols (e.g., tokens) derived from Turing and Von Neumann machines.
It typically means fixed serial rules with variables, e.g., computer programs.
The Physical Symbol System (PSS) hypothesis~\cite{newel1976computer} was developed to study the systematic mental representations of humans, and it says that human cognition is physically the product of a symbol system\footnote{``A physical symbol system has the necessary and sufficient means for intelligent action''~\cite{newel1976computer}.}.
Connectionist~\cite{feldman1982connectionist,rumelhart1986parallel} uses many simple neuron-like units which are richly interconnected and processed in parallel~\cite{1991Connectionist}.
It implies learning from data without explicit symbol information, e.g., neural networks.

While Connectionist is good at i.i.d. generalization, it does not address o.o.d. generalization in general~\cite{fodor1988connectionism}.
On the other hand, Classicist is good at o.o.d. generalization, while not at i.i.d. generalization (sometimes referred to as the ``graceful degradation'' issue).
For example, computer programs can be reliable for new data that strictly fit input requirements, but it is brittle to process noisy images or speeches correctly.
There is a spectrum between Classicist and Connectionist, which is a trade-off between the advantages of both approaches (Section~\ref{sec:types_of_connectionist}).

\paragraph{Distributed representation}
Connectionist has distributed and localist representations~\cite{feldman1986neural,sejnowski1987parallel}.
We mainly discuss distributed representation, which is widely used and is more efficient when many items are present~\cite{touretzky1988distributed}.
It refers to Parallel Distributed Processing (PDP) models~\cite{rumelhart1986parallel}.

Distributed representations of symbols~\cite{rumelhart1986Learning} were introduced to capture relationships between family members.
A distributed representation can describe an object in terms of primitive descriptors; it has a significant advantage because it can describe a novel object using the same primitive descriptors to create novel combinations as representations~\cite{hinton1990mapping}.
During training, primitive descriptors or hidden nodes in distributed representation continually change their meanings; though they might be stable in the short term, they shift around in the longer term~\cite{hinton1986learning}.

Localist representation often refers to a one-hot representation when used in deep learning.
It is efficient in the following cases~\cite{hinton1990mapping}.
A significant portion of samples activates a node, e.g., the end-of-sentence symbol in sentences.
Entries are mutually exclusive, e.g., classification outputs and input words.

\paragraph{Disentangled representation}
Distributed representation can be disentangled~\cite{bengio2013representation}.
With fewer requirements, e.g., linearity, it is also referred to as factorized representation~\cite{ke2021systematic} or microfeatures~\cite{Hinton1986Distributed}.
Many early works for systematic generalization studies disentangled representation.
Also, disentangled representation learning is mainly studied in unsupervised manners~\cite{higgins2017beta}, and it can be used as a feature extractor for systematic generalization models.
These previous works usually do not discuss disentangled representation learning and systematic generalization together.
However, it is necessary in some cases because humans systematically generalize many entangled data, such as images and sentences.

Disentangled representation is also related to Invariant Risk Minimization (IRM)~\cite{InvariantRiskMinimization,peters2016causal}, a learning paradigm that estimates invariant predictors from multiple training environments.
IRM focuses on learning individual features.
However, disentangled representations require decomposing an input to features, and the features together should keep the original input information~\cite{bengio2013representation}.
IRM is used in domain generalization, and there is a recent related survey~\cite{wang2022generalizing}.

\paragraph{Connectionist and systematic generalization}
It is argued that Connectionism lacks systematicity, which also indicates that the mind is not a Connectionist network~\cite{fodor1988connectionism}.
Current machine learning methods also seem weak at generalization beyond the training distribution~\cite{goyal2021recurrent,hendrycks2018benchmarking}, though it is often necessary in many cases.
Also, state-of-the-art models often learn spurious statistical patterns while humans avoid them~\cite{nie-etal-2020-adversarial}.

There has been the exploration of compositionality in neural networks for systematic behavior~\cite{wong2007generalisation,brakel2009strong}, counting ability~\cite{rodriguez1998recurrent,weiss-etal-2018-practical}, and sensitivity to the hierarchical structure~\cite{linzen2016assessing}.
Systematicity has been partially achieved in previous work~\cite{Niklasson94canconnectionist,hadley1997strong,boden2000semantic}.
More recent related efforts are summarized~\cite{jansen2012strong}.
There are also recent improvements in systematic generalization with structure design~\cite{andreas2016neural,gaunt2017differentiable} and structure prediction~\cite{johnson2017inferring,hu2017learning}.
More recently, neural production systems~\cite{goyal2021factorizing,goyal2021neural}, shared global workspace~\cite{goyal2022coordination}, and reinforcement learning~\cite{ke2021systematic} are also investigated to address this problem.
We will discuss more in Section~\ref{sec:recent_improvements}.

\paragraph{System 1 and System 2 in human thinking}
Human thinking has two systems.
System 1 refers to fast and less reliable thinking.
People always make unconscious decisions, and the process is mundane reasoning.
System 2 refers to slow and more reliable thinking.
It is a conscious process including logical thinking, e.g., solving a math problem.
System 2 is a more precious resource.
People switch between the two systems to best use the resources.
We mainly use system 1 in daily life and system 2 for complicated problems that system 1 cannot address.
An introductory book~\cite{kahneman2011thinking} contains many examples, experiments, and discussions of how humans think on different occasions.

Systematic generalization is more related to system 2, where a decision in a new environment is inferred with logical reasoning with familiar knowledge.
Human thinking is also discussed from other perspectives.
Dynamic memory is considered a fundamental ingredient of intelligence~\cite{Schank1982DynamicM}.
Consciousness prior~\cite{bengio2017consciousness} is also proposed in deep learning.
It leads to sparseness prior and attention-based modular networks.

\section{Types of Connectionist}
\label{sec:types_of_connectionist}
Classicists and Connectionists have complementary advantages.
Implementational Connectionist merely implements a Classicist symbol manipulation system.
On the contrary, eliminative or radical Connectionist is not designed with knowledge of the symbol system, so it eliminates the PSS hypothesis.
Also, there are hybrid methods that combine both of them.
Eliminative, hybrid, and implementational Connectionists have weak to intense use of symbol systems.

\subsection{Eliminative Connectionist}
Eliminative Connectionist eliminates the symbol system by purely using distributed representation~\cite{pinker1988language,marcus1998rethinking}.
It is attractive because it avoids knowledge of symbol systems.
For example, a model without symbolic rules or lexicon can learn past tense grammar rules and exceptions in English verbs~\cite{rumelhart1986on}.

Eliminative Connectionist also uses prior knowledge, not including symbol systems, e.g., layered architecture.
Also, some architecture designs, such as convolutional layers and attention mechanisms, are less mentioned as symbol systems.
Regularization algorithms like noise insertion and optimization algorithms like Adam are not much related to symbol systems.
Modular architecture design a module for a factor, but the prior knowledge is not specific to each symbol.

Pure eliminative Connectionist has many challenges.
One long-standing problem is that a model has a fixed-length vector of units for representations of recursive symbol structure~\cite{pollack1990recursive}, which vary in size and complexity~\cite{Holyoak2000ThePT}, e.g., context-free languages.
Tree-structured composition~\cite{bowman2015treestructured} is developed to address such cases.

\subsection{Implementational Connectionist}
Implementational Connectionist~\cite{ballard1986cortical,pinker1988language,Hinton1981Implementing,Hinton1986Distributed,1986BoltzCONS} directly designs symbol-processing architectures in PDP models~\cite{chalmers1993connectionism}.
It was argued that the only Connectionist approach to achieve systematic generalization is implementational Connectionist~\cite{fodor1988connectionism}.
Such approaches share the explanatory capability and the empirical results of Classicist models.

For example, PDP networks can implement parts of LISP and production systems~\cite{touretzky1985symbols}.
BoltzCONS~\cite{1986BoltzCONS} is another example.
Also, $\mu$KLONE~\cite{derthick1990muldane} uses microfeatures to implement functionality similar to the knowledge representation system KL-ONE.

\subsection{Hybrid Neural Systems}
It is proposed to integrate symbolic processing into neural networks~\cite{1990Preface,Sun1996Hybrid,wermter1998an} to take advantage of both.
The majority of early hybrid systems have a neural network and rule-based modules~\cite{mcgarry1999hybrid}.

Like classicists, the neural net was designed to separate modules for storing values and operations~\cite{miikkulainen1993subsymbolic}.
It has been shown that some logic can be translated into neural networks~\cite{shavlik1994combining}.
Methods are proposed for variable binding (Section~\ref{sec:variable_binding}) with tensor products~\cite{smolensky1990tensor} and semantic pointers~\cite{eliasmith2013build}.
There is also semi-local tensor product representation, such as the semantic net model~\cite{Hinton1981Implementing}.
Neural-symbolic methods are proposed for logical operations~\cite{Niklasson94canconnectionist} and reasoning~\cite{2009nscr.bookD,garcez2019neuralsymbolic}.

Recently, the Neuro-Symbolic concept learner~\cite{mao2019the} has been proposed for VQA.
Tensor-Product Transformer combines BERT and tensor products to represent symbolic variables and their bindings~\cite{schlag2020enhancing}.
Neural-symbol stack machines~\cite{NEURIPS2020_12b1e42d} are used for instruction learning problems.
Recent work also includes Edge Transformers~\cite{bergen2021systematic}, which combines Transformers and rule-based symbolic systems.
Inspired by logical programming, it proposes triangular attention to manipulate pairs of input nodes.
It is proposed to use symbolic modules to examine the logical reasoning of neural sequence modules~\cite{nye2021improving}.
A book~\cite{1991Connectionist} contains works for connectionist symbol processing.

\section{Variable binding}
\label{sec:variable_binding}
Variable binding assigns a value to a variable.
It is a difficult and important problem in Connectionist models~\cite{Barnden1984On}, and it is required for complex reasoning tasks~\cite{Browne00connectionistvariable} and more efficient computation~\cite{sun1992on}.
Systematic generalization requires learning true rules containing variables, so it must do something equivalent to the variable binding~\cite{touretzky1988distributed}.
Manipulation of variables is essential for animal cognition~\cite{2009Memory}.
For example, honeybees extend the solar azimuth function to the lighting of unseen conditions~\cite{Dyer1994Development}.
Variable binding may be one of the reasons to force people to be sequential processors~\cite{newell1980harpy}.

We look at an example.
Variable binding enables one general rule: dog(X), barks(X). It means ``X barks if X is a dog.''
Without variable binding, we need a specific rule for each possible value, such as dog(rei) and dog(bue)~\cite{Browne00connectionistvariable}.
Similarly, the binding problem occurs when we encode feature conjunctions in a representation, e.g., a red triangle and a blue square~\cite{treisman1998feature}.
Production systems~\cite{touretzky1988distributed,goyal2021neural} have rule and placeholder variables, and variable binding is required for both of them.

It is argued that eliminative Connectionists cannot explicitly address the variable binding problem~\cite{Holyoak2000ThePT}.
Many Connectionist researchers have considered embedding symbol systems in a neural network for variable binding~\cite{feldman1982connectionist,pollack1990recursive}.
Holographic Reduced Representations~\cite{Plate91holographicreduced} use convolution. Temporal Synchrony~\cite{1993From} is proposed for reasoning. Analogical Access and Mapping~\cite{e33d4f80b03d4ce5ac2807e9ff3647f7} mainly regards tree-structure grammar in language.
Please also refer to recent surveys~\cite{10.1162/neco_a_01179,frady2021variable}.

\section{Causality}
\label{sec:causality}
Causal learning has a long history, rooted in the eighteenth century~\citep{hume2003treatise} and the classical field of AI~\citep{pearl2003causality}.
The primary exploration has been from statistical perspectives~\citep{pearl2009causality,peters2016causal,greenland1999causal,pearl2018does}.
Much causal reasoning literature is built upon do-calculus~\citep{pearl1995causal,pearl2009causality} and interventions~\citep{peters2016causal}, though some early work does not consider interventions~\citep{heckerman1995learning}.
The question of how to separate correlation and causation is raised~\citep{welling2015ml}.

The causation forms Independent Causal Mechanisms~\cite{peters2017elements,scholkopf2021toward}, or ICMs, which avoid spurious connections\footnote{``ICM principle: The causal generative process of a system's variables is composed of autonomous modules that do not inform or influence each other''~\cite{scholkopf2021toward}.}.
ICM is robust across different domains~\cite{SchJanLop16} to support systematic generalization~\cite{parascandolo2018learning,goyal2021recurrent}.

Causality and variable binding have been discussed in different fields while closely related.
For example, an ICM states that an output variable depends only on a corresponding input variable.
It binds the output variable and the values of a factor in a disentangled input variable.
Causality also indicates that the binding is robust in different domains.

Systematic generalization is the counterfactual when the joint input distribution is intervened to have new values with zero probability in training (covariate shift).
With ICMs, models can be trained with data distributions induced by causal models to achieve systematic generalization~\cite{tsirtsis2020optimal}.
For example, causal mechanisms are augmented into generative models for constructing images and planning~\citep{kocaoglu2018causalgan,kurutach2018learning}.

As mentioned in the book~\cite{peters2017elements}, causality research is still in an early stage, and the assumptions are not general.
The theory has more results on linear models.
Some main approaches include the following.
\begin{itemize}
    \item Independence-based methods
    \item Restricted structural models, such as additive noise
    \item Invariant causal prediction
\end{itemize}

The current approaches mainly study disentangled input representations.
Research for entangled representations has been recently conducted~\cite{bengio2020a}, and it uses adaptation speed to learn causality with meta-learning.
However, extending such a representation-learning algorithm to multiple variables is still challenging.
Some work includes multiple variables but disentangled data~\cite{ke2019learning}.

One possibility is to use the causality module as an intermediate part of a neural network model~\cite{scholkopf2021toward}.
A model can be divided into an encoder, an intermediate network, and a decoder.
The encoder converts an entangled input to a disentangled input representation.
The intermediate network models the causality between input and output disentangled representations.
The decoder converts a disentangled output representation to an entangled output.
Graph networks~\cite{battaglia2018relational} can be used as intermediate networks~\cite{ke2021systematic}.

\section{Language, Vision, and VQA}
\label{sec:language_vision_vqa}
Systematic generalization has been studied in different fields, and we look into language, vision, and VQA.

\subsection{Language}
Systematicity is often referred to as compositionality in the language domain.
They may be different aspects of the same phenomenon~\cite{fodor1988connectionism}.
Compositionality is the algebraic capacity to understand and produce novel combinations from known components~\cite{chomsky1957syntactic,montague1970universal}.
For example, a person who knows how to ``step,'' ``step twice,'' and ``jump'' naturally knows how to ``jump twice''~\cite{lake2018generalization}.
This generalization ability is critical in human cognition~\cite{minsky1986society,lake2017building}.
It helps humans to learn languages flexibly and efficiently from limited data and extend to unseen sentences.

Human-level compositional learning has been an open challenge~\cite{yang2019task}.
With the breakthroughs in sequence-to-sequence neural networks for NLP, such as RNN~\cite{sutskever2014sequence}, Attention~\cite{xu2015show}, Pointer Network~\cite{vinyals2015pointer}, and Transformer~\cite{vaswani2017attention}, there are more contemporary attempts to encode compositionality in sequence-to-sequence neural networks.
Words are natural symbols in language and are extended to word embeddings~\cite{Deerwester90indexingby}.
Further, neural language models~\cite{bengio2003A} introduce interpretable word embeddings.

SCAN dataset~\cite{lake2018generalization} is an early compositional generalization dataset in recent years.
It is a sequence-to-sequence task that translates natural language into a sequence of robot actions.
It considers several aspects of compositional generalization.
One of them is primitive substitutions, where a word is replaced with another, and the combination of the word and the context is new.
Please see the ``jump'' example above.
Many related tasks~\cite{loula-etal-2018-rearranging,livska2018memorize,bastings-etal-2018-jump,lake2019human} are also proposed.

Multiple methods~\cite{bastings-etal-2018-jump,loula-etal-2018-rearranging,kliegl2018more,chang2018automatically} have been proposed using various RNN models and attention mechanisms.
These methods successfully generalize when the difference between training and test data is slight.
Requirements for systematic generalization are discussed~\cite{bahdanau2018systematic}, concluding that additional regularization or prior is necessary for modular designs.
SCAN dataset inspired multiple approaches~\citep{russin2019compositional,lake2019compositional,li2019compositional,andreas-2020-good,Gordon2020Permutation,liu2020compositional,NEURIPS2020_12b1e42d} discussed in the next section.

The CFQ dataset considers syntactic compositionality in real data~\cite{keysers2020measuring}.
It generally requires recombining syntactic structures beyond primitive substitution.
The methods on the SCAN dataset do not work well on CFQ, while pretraining provides improvements~\cite{furrer2020compositional}.
The Semantic Parsing approach also addresses part of the problem~\cite{shaw2021compositional}.
There are analyses for training data size~\cite{tsarkov2021cfq} and model size~\cite{qiu2022evaluating} for compositional generalization.

There are other recent semantic parsing datasets.
COGS~\cite{kim-linzen-2020-cogs} is a synthetic dataset with pairs of sentences and logical forms, and the generalization test set evaluates novel linguistic structures.
PCFG~\cite{ijcai2020-708} manipulates executable operations.
GeoQuery~\cite{shaw2021compositional} is a non-synthetic dataset with pairs of questions and meaning representations annotated by humans.
It has three systematic generalization splits.
Template split has disjoint abstract output templates for training and test data.
TMCD split has training and test compound distributions as divergent as possible.
Length split has different lengths for training and test data.
SMCalFlow-CS~\cite{yin-etal-2021-compositional} is a split of SMCalFlow for compositional skills.
Machine translation dataset is also recently proposed~\cite{dankers-etal-2022-paradox}.
Math expressions can be treated as language, and a mathematical reasoning dataset is proposed~\cite{saxton2018analysing}.

\subsection{Vision}
Systematic generalization is often referred to as zero-shot learning in vision~\cite{rohrbach2011evaluating,larochelle2008zero,yu2010attribute,xu2017matrix,ding2017low}.
The difference is that vision tasks are additionally given attributes (factors) for classes or samples.
Common datasets include AWA~\cite{lampert2014attribute,xian2019zero}, CUB~\cite{wah2011the}, SUN~\cite{patterson2012sun}, and aPY~\cite{farhadi2009describing}.
There are also recent vision benchmarks~\cite{hendrycks2018benchmarking,hendrycks2020augmix,tang2021crossnorm} for systematic generalization.

Many approaches have been proposed with linear~\cite{frome2013devise,romera-paredes2015an,akata2013label,akata2015evaluation} and nonlinear~\cite{socher2013zero,norouzi2014zero} compatibility models.
Other algorithms learn independent attributes~\cite{lampert2014attribute}.
There are also hybrid models between them~\cite{changpinyo2016synthesized,zhang2015zero,xian2016latent}.
There are related surveys~\cite{WANG2018135,zhou2022domain}.

Using attributes or other side information makes the problem easier than systematic generalization.
Many works have been done to avoid attribute annotation, e.g., one-shot image novel class~\cite{mensink2012metric}, external lexical information for class embeddings~\cite{rohrbach2011evaluating,akata2015evaluation}, and visual descriptions~\cite{reed2016learning}.
Other work has been done to understand the systematicity of images~\cite{goyal2022coordination}.
It has also been argued that zero-shot learning is related to the attention mechanism~\cite{sylvain2020locality}.

A related topic is domain generalization with multiple vision datasets, such as PACS~\cite{li2017deeper}, VLCS~\cite{torralba2011unbiased}, MNIST-M~\cite{ganin2015unsupervised}, and NICO~\cite{he2021towards,zhang2022nico++}. 
NICO labels both concept and context and the context can be attributes or backgrounds.

\subsection{VQA}
Both language and vision are essential for human recognition, and VQA~\cite{antol2015vqa} combines them.
VQA naturally includes grounding, which finds the mapping between words and objects or their properties.
Systematicity is also applicable and critical in other multimodal problems, including Image Captioning~\cite{karpathy2015deep}, Image Generation~\cite{klinger2020study}, and Embodied Question Answering~\cite{das2018embodied}.

In early VQA, it was found that the trained models are likely to learn superficial and spurious relations between input and output.
For example, when a question asks what is on the ground, the answer is likely to be snow.
It is because the snow on the ground is worth to be asked.
They are systematic generalization problems.

VQA datasets are designed for systematic generalization.
CLEVR~\cite{johnson2017clevr} contains Compositional Generalization Test (CoGenT) for novel attribute combinations in the test.
CLOSURE~\cite{bahdanau2019closure} measures systematic generalization in the CLEVR dataset.
Another VQA dataset is SQOOP~\cite{bahdanau2018systematic}.
GQA is a more realistic dataset~\cite{hudson2019gqa}.

Algorithms for visual question answering include architecture design of Neural Module Networks~\cite{andreas2016neural}, Film~\cite{perez2018film}, Relation Networks~\cite{santoro2017simple}, and MAC networks~\cite{hudson2018compositional}.
Latent Compositional Representation~\cite{bogin2021latent} also helps.

Also, following the SCAN dataset for one-shot learning in language, the gSCAN dataset was proposed for one-shot learning problems in grounding and visual question answering~\cite{ruis2020benchmark}.
The input is a human language instruction and an environment, and the output is a sequence of robot actions.
A study on gSCAN shows that it is crucial to think before acting~\cite{heinze2020think}.
The object relations are modeled in the contexts~\cite{gao-etal-2020-systematic}.
It is important to fit the network structure to the compositional structure of the problem~\cite{kuo-etal-2021-compositional-networks}.
A general transformer with cross-modal attention achieves nearly perfect results for majority splits, and the remaining problems correspond to the fundamental challenges of compositional generalization for language~\cite{qiu-etal-2021-systematic}.

There are also various simulated settings for grounded language acquisition with reinforcement learning, such as X World~\cite{yu2018interactive}, BabyAI~\cite{chevalier-boisvert2018babyai}, and others~\cite{hermann2017grounded,wu2018building}.

\section{Recent Improvements}
\label{sec:recent_improvements}
Different systematic generalization approaches have been investigated.
However, the generalization is still difficult for deep learning in general~\cite{hendrycks2018benchmarking,goyal2021recurrent}.
The main directions include disentangled representation learning, meta-learning, attention mechanism, modular architectures, specialized architectures, and data augmentation.

\subsection{Disentangled Representation Learning}
It was argued that good representations should help express the regularities~\cite{hinton1990mapping}.
Disentangled representation~\cite{bengio2013representation} learning is developing quickly.
Early work learns the representation from statistical marginal independence~\cite{higgins2017beta,burgess2018understanding,locatello2019challenging}.

The definition of disentangled representation has recently been proposed with symmetry transformation in Physics~\cite{higgins2018towards}.
It leads to Symmetry-based Disentangled Representation Learning~\cite{NEURIPS2019_36e729ec,NEURIPS2020_e449b931,NEURIPS2020_9a02387b,NEURIPS2020_c9f029a6}.
Such approaches explain disentangled representation using group theory and Physics.

It is mentioned that disentangled representation is an example of ICM learning~\cite{scholkopf2021toward}.
There are also methods to measure compositionality in representations~\cite{andreas2018measuring}.
Disentangled representation tends to be discussed without simultaneous systematic generalization.
It can be a feature extractor to obtain disentangled representations, and in other systematic generalization tasks, the representations are used as inputs for downstream modules.

\subsection{Meta-learning}
Meta-learning is an approach for systematic generalization~\cite{lake2019compositional}.
It usually designs a series of training tasks for learning a meta-learner, which is used to address the problem in the target task.
There is training and test data in each training task, where test data requires systematic generalization from training data.
The training tasks need to have similar structures as the target task so that the meta-learner can learn how to generalize from the training data in the target task.

When ICMs are available, they can be used to generate meta-learning tasks~\cite{scholkopf2021toward}.
It is discussed to employ meta-reinforcement learning for causal reasoning~\cite{dasgupta2019causal}.
Meta-learning can also capture the adaptation speed to discover causal relations~\cite{bengio2020a,ke2019learning}.
However, it is hard to disentangle the factors when multiple variables exist.

There are other works with meta-learning.
Pairs of meta-learning tasks are constructed from sub-sampling training data~\cite{conklin-etal-2021-meta}.
Representation and task-specific layers of models are trained differently
to generalize mismatched splits on pre-finetuning tasks, so transfer learning between compositional generalization tasks is enabled~\cite{zhu2021learning}.

\subsection{Attention Mechanism}
Attention mechanisms, especially key-value attention mechanisms, are widely used in neural networks~\cite{bahdanau2015neural}.
The key-value mechanisms are composed of a query, keys, and values.
The query and the keys generate an attention map, which extracts a value from the values.
An attention map is similar to a pointer, often used in symbol processing.
It is also a type of distal access~\cite{Newell1980Physical}, which uses an abbreviated tag for referring to a structure.
A symbol is informally regarded as a small representation of an object, which provides ``remote access'' for the fuller representation of an object~\cite{hinton1990mapping}.

Transformers~\cite{vaswani2017attention} are modern neural network architectures with self-attention.
Recurrent Independent Mechanisms~\cite{goyal2021recurrent} use attention mechanisms and the names of the incoming nodes for variable binding.
Global workspace~\cite{goyal2022coordination} improves them by using limited-capacity global communication to enable the exchangeability of knowledge for systematic generalization.
Discrete-valued communication bottleneck~\cite{liu2021discrete} further enhances the generalization.

Different extensions to attention modules are discussed~\cite{oren-etal-2020-improving}.
Auxiliary objectives to bias attention in encoder-decoder models are proposed~\cite{yin-etal-2021-compositional,jiang2021inducing}.
There are also sparse variants~\cite{shazeer2017} of attention.
Compositional Attention~\cite{mittal2022compositional} disentangles search and retrieval in Transformer architecture.
It addresses redundancies in multi-head attention with different numbers of searches and retrievals and dynamic selection.

We like to discuss the relationship between the attention mechanism and ICMs.
The sparse connection prior knowledge~\cite{bengio2017consciousness} has two types.
The first is the sparseness on a dynamic graph or the routes for each sample.
It corresponds to attention mechanisms.
The second is the sparseness on a static graph or the connections between variables.
It corresponds to ICMs.
ICMs enable systematic generalization.
The dynamic sparseness may not infer the static one, so attention may not establish ICMs to enable the generalization.
However, attention reduces the size of a module input, so test inputs are more likely to remain in the training domain, which helps systematic generalization.
For example, a word is a part of an input sentence, so an attended word can be correctly processed, even if the sentence is unseen.
Also, the attention mechanism is usually an operator and does not contain parameters, so the mechanism suffers less from the change of distribution.

\subsection{Modular architectures}
Modular architecture has a long history, such as the mixture of experts~\cite{jacobs1991adaptive,jordan1994hierarchical}.
Early related ideas apply micro-inference, which uses some of the features of some of the role-fillers to infer some of the features of the other role-fillers~\cite{hinton1990mapping}.
There are also recent results~\cite{graves2014neural,andreas2016neural,hu2017learning,vaswani2017attention,goyal2021recurrent,goyal2021neural,mittal2020learning,ke2021systematic}.

Modular architectures are natural for combinatorial generalization~\cite{battaglia2018relational}.
There are task-specific modular networks~\cite{jacobs1991task}.
Though modules can be designed for different factors, the input to each module may still have spurious influence from other factors when the model input is entangled.
It can be helpful to regularize entropy to bottleneck modules in such cases~\cite{li2019compositional}.

Attention mechanisms can be used with modular architecture~\cite{riemer2016correcting,peters2017elements,mittal2022modular}.
Object-centric slot attention~\cite{locatello2020object} finds objects for downstream networks.
Neural Interpreters~\cite{rahaman2021dynamic} factorize inferences to modules in a self-attention network.
It can be trained end-to-end by routing through modules.

Modular and compositional computation~\cite{rosenbaum2019routing} in routing networks were analyzed.
A differentiable weight mask is used to examine the modularity of neural networks.
It finds that neural networks are not trained to be modular.
Common modular architectures are assessed~\cite{mittal2022modular} with collapse and specialization problems, finding end-to-end learned modular systems are not optimal.

\subsection{Specialized architectures}
Another common approach is specialized architecture design~\cite{russin2019compositional,Gordon2020Permutation,liu2020compositional,NEURIPS2020_12b1e42d}.
The importance of design decisions is reported~\cite{ontanon-etal-2022-making}.

Transformers significantly improve semantic parsing when model configurations are carefully adjusted, and Universal Transformer variants also work well~\cite{csordas-etal-2021-devil}.
Reordering and aligning the structure~\cite{NEURIPS2021_6f46dd17} can model segment-to-segment alignments with a neural reordering module for separable permutations.
The span-based parser~\cite{herzig-berant-2021-span} treats a tree as a hidden variable.

Large pre-trained language models convert inputs to intermediate language representations for semantic parsing~\cite{shin-etal-2021-constrained}.
Intermediate representation helps compositional generalization for pre-trained seq2seq models~\cite{herzig2021unlocking}.
Program synthesis~\cite{nye2020learning} learns explicit programs from training data.
Semantic tagging~\cite{zheng-lapata-2021-compositional-generalization} trains an alignment tagger by entity linking with $\lambda$-calculus and SQL expressions.
It uses tags to supervise hidden variables.
Iterative decoding~\cite{ruiz2021iterative} breaks training examples down into a sequence of intermediate steps.

\subsection{Data augmentation}
Data augmentation is primarily for language tasks, as words and phrases in a sentence are more straightforward to modify than the pixels in images.
Multiple approaches are proposed for adding data~\cite{guo2021revisiting,wang2021learning,guo-etal-2020-sequence} and training from labeled data~\cite{yu2021grappa,zhong-etal-2020-grounded}.

GECA (good-enough compositional augmentation)~\cite{andreas-2020-good} is a rule-based protocol for sequence modeling.
It provides inductive bias for compositionality.
It can replace discontinuous sentence fragments, e.g., ``\textit{Tom \underline{picks} apples \underline{up}}.''
R\&R (recombine and resample)~\cite{akyurek2021learning} learns schemes for data augmentation.
It replaces the symbolic generative process with neural models and obtains the inductive bias as explicit rules.

Data recombination~\cite{jia2016data} injects task-specific prior knowledge for modeling logical regularities in semantic parsing.
It induces synchronous context-free grammar from training data.
It is also studied that the diverse sampling structure of synthetic examples helps systematic generalization~\cite{oren-etal-2021-finding}.
CSL (Compositional Structure Learning)~\cite{qiu-etal-2022-improving} is a generative model with context-free grammar induced from training data.
The examples from CSL are recombined and used to fine-tune a pre-trained model.
It is studied to use subtree substitution~\cite{yang2022subs} for data augmentation.

There is also data augmentation for images, e.g., interpolating both image input and label output~\cite{pmlr-v162-yao22b}.
Stable learning~\cite{9577393} learns weights for training samples to remove dependencies between features.

\section{Conclusion}
Systematic generalization is a critical capability for artificial intelligence.
While it is straightforward for classic symbol processing approaches, it is difficult for Connectionist approaches.
It has been discussed with crucial problems of variable binding and causal learning.
Our discussion covers different AI fields, such as language, vision, and VQA, and there are recent improvements in different aspects.
Though some specific problems are addressed, there are still many things unknown about systematic generalization in deep learning.
We hope this survey helps in understanding the background and inspiring future work.

\section*{Acknowledgments}
We thank Liang Zhao for beneficial discussions, suggestions, and adding information.
We also thank Yi Yang for the helpful advice.

\bibliography{main}

\end{document}